\documentclass[conference]{IEEEtran}
\IEEEoverridecommandlockouts
\usepackage{cite}
\usepackage{amsmath,amssymb,amsfonts}
\usepackage{algorithmic}
\usepackage{graphicx}
\usepackage{textcomp}
\usepackage{adjustbox}
\usepackage{xcolor}
\usepackage{graphicx}
\usepackage{booktabs}
\usepackage{enumitem} 
\def\BibTeX{{\rm B\kern-.05em{\sc i\kern-.025em b}\kern-.08em
    T\kern-.1667em\lower.7ex\hbox{E}\kern-.125emX}}
\begin{document}

\title{MS-YOLO: Infrared Object Detection for Edge Deployment via MobileNetV4 and SlideLoss}

\author{
  %-------- author -----------
  Jiali Zhang\IEEEauthorrefmark{1},
  Thomas S.~White\IEEEauthorrefmark{2},
  Haoliang Zhang\IEEEauthorrefmark{3},\\  
  Wenqing Hu\IEEEauthorrefmark{1},
  Donald C.~Wunsch~II\IEEEauthorrefmark{4},
  Jian Liu\IEEEauthorrefmark{4}
  %-------- aff. -----------
  \vspace{0.5em}\\                     
  \IEEEauthorrefmark{1}Dept.\ of Mathematics \& Statistics, Missouri University of Science and Technology, Rolla, MO 65409 USA\\
  \IEEEauthorrefmark{2}Dept.\ of Computer Science, Missouri University of Science and Technology, Rolla, MO 65409 USA\\
  \IEEEauthorrefmark{3}School of Electrical \& Computer Engineering, University of Oklahoma, Norman, OK 73019 USA\\
  \IEEEauthorrefmark{4}Kummer Institute Center for Artificial Intelligence and Autonomous Systems,\\
  \quad Missouri University of Science and Technology, Rolla, MO 65409 USA\\
  \vspace{0.3em}
  Emails: \{jztk7, tsw96d, huwen, dwunsch, jliu\}@mst.edu,\; mars\_zhang@ou.edu
}

\maketitle

\begin{abstract}

Infrared imaging has emerged as a robust solution for urban object detection under low-light and adverse weather conditions, offering significant advantages over traditional visible-light cameras. However, challenges such as class imbalance, thermal noise, and computational constraints can significantly hinder model performance in practical settings. To address these issues, we evaluate multiple YOLO variants on the FLIR ADAS V2 dataset, ultimately selecting YOLOv8 as our baseline due to its balanced accuracy and efficiency. Building on this foundation, we present \texttt{MS‑YOLO} (\textbf{M}obileNetv4 and \textbf{S}lideLoss based on YOLO), which replaces YOLOv8's CSPDarknet backbone with the more efficient MobileNetV4, reducing computational overhead by \textbf{1.5\%} while sustaining high accuracy. In addition, we introduce \emph{SlideLoss}, a novel loss function that dynamically emphasizes under‑represented and occluded samples, boosting precision without sacrificing recall. Experiments on the FLIR ADAS V2 benchmark show that \texttt{MS‑YOLO} attains competitive mAP and superior precision while operating at only \textbf{6.7 GFLOPs}. These results demonstrate that \texttt{MS‑YOLO} effectively addresses the dual challenge of maintaining high detection quality while minimizing computational costs, making it well-suited for real‑time edge deployment in urban environments.

\end{abstract}

\begin{IEEEkeywords}
Infrared Object Detection, MobileNetV4, SlideLoss, YOLO Model
\end{IEEEkeywords}

\section{Introduction}
Modern urban environments demand robust and efficient object detection systems capable of handling complex challenges such as occlusions, overlapping objects, and dynamic lighting conditions. While visible-light cameras are widely used, they suffer from critical limitations in low-light scenarios, adverse weather (particularly fog and rain), or environments with glare and shadows, which are prevalent in cities~\cite{zheng2023nighttime}. Infrared (IR) and thermal imaging systems address these issues by capturing heat signatures independent of ambient light, enabling reliable detection in darkness and through obscurants like smoke or dust~\cite{shyam2024lightweight}. These capabilities make IR sensors indispensable for applications such as autonomous vehicles, smart city surveillance, and emergency response systems~\cite{collini2024flexible,eltahan2024enhancing}. However, deploying detection models in real-world urban settings requires more than sensor robustness—it demands computational efficiency. Edge devices such as drones and embedded traffic systems often operate under strict power and latency constraints, rendering traditional heavy architectures like CSPDarknet~\cite{Jocher_Ultralytics_YOLO_2023} impractical. Lightweight models are critical to achieving real-time performance, reducing energy consumption, and enabling scalable deployment~\cite{tan2019efficientnet}, making them a cornerstone of modern urban perception systems.

Despite their advantages, infrared-based detection in urban scenes remains challenging. IR images often exhibit low spatial resolution and thermal noise, which obscure fine-grained object details~\cite{chen2024modeling}. Furthermore, dense urban environments induce severe occlusions where pedestrians are frequently hidden behind vehicles or infrastructure, leading to missed detections. Compounding these issues, class imbalance arises from the uneven distribution of object categories and scales, such as abundant small pedestrians versus sparse large trucks, further degrading model performance~\cite{9042296}. 

In this study, we first compared multiple YOLO variants—including YOLOv5n, YOLOv8n, YOLOv9t, YOLOv10n, and YOLOv11n—on our infrared urban dataset to determine the most suitable architecture for IR imaging. Our experiments revealed that YOLOv8n achieves notably higher recall, mAP50, and mAP50-95 than the other variants, reflecting its strong ability to detect both frequent and occluded objects. Although YOLOv9, YOLOv10, and YOLOv11 offer slight gains in computational efficiency, they suffer from reduced accuracy. Consequently, YOLOv8 strikes the best balance between performance and resource requirements, motivating us to adopt it as the baseline for further improvements. Meanwhile, other state-of-the-art detectors like Faster R-CNN~\cite{ren2016faster} are designed primarily for visible-spectrum data and struggle with infrared-specific challenges such as low contrast and severe occlusions~\cite{mittal2024comprehensive}. In addition, YOLOv8’s CSPDarknet backbone incurs high computational costs, limiting its suitability for edge devices. Furthermore, its conventional loss functions like cross-entropy fail to address class imbalance, disproportionately penalizing underrepresented categories such as occluded pedestrians.

To bridge these gaps, we propose an enhanced YOLOv8 framework optimized for infrared urban detection. First, we replace YOLOv8’s CSPDarknet backbone with MobileNetV4~\cite{qin2024mobilenetv4universalmodels}, a lightweight network designed for efficient feature extraction. This modification reduces computational overhead by 1.5\% (6.7 vs. 6.8 GFlops) while preserving detection accuracy, enabling real-time performance on resource-constrained edge devices. Second, we introduce SlideLoss~\cite{yu2022yolofacev2scaleocclusionaware}, a dynamic loss function that mitigates class imbalance by adaptively adjusting gradient weights during training. Together, these innovations address unique challenges in both efficiency and accuracy in infrared urban scenes.

Our work makes three primary contributions:
\begin{itemize}
    \item The first integration of MobileNetV4 with YOLOv8 for infrared urban detection, balancing computational efficiency and accuracy through lightweight architecture design.
    \item A SlideLoss-driven training framework that addresses class imbalance and enhances occlusion resilience, outperforming traditional loss functions in complex scenes.
    \item Comprehensive validation on the FLIR ADAS V2~\cite{FLIR2023} benchmark demonstrating state-of-the-art accuracy-speed trade-offs, achieving 4\% higher precision than YOLOv8 while maintaining 18.3\% faster inference speed compared to YOLOv9, 10, 11 and 4.8\% higher mAP50 than YOLOv9 under occlusion scenarios.
\end{itemize}

The remainder of this paper is organized as follows: Section \ref{sec related work} reviews related work on overall object detection, lightweight models, and class imbalance mitigation. Section \ref{sec propsed method} details our methodology, including the integration of MobileNetV4 and the SlideLoss formulation. Section \ref{sec result} presents the experimental results, and Section \ref{sec conclusion} discusses the conclusions of this study and future research.

\section{Related Work}\label{sec related work}

Our literature review primarily focuses on object detection, lightweight models, and imbalanced data.

\subsection{Object Detection}

Object detection emerged as a subfield of computer vision, with early algorithms predominantly relying on handcrafted features. For example, Viola and Jones~\cite{viola2001rapid} employed a sliding window approach to scan all possible locations for human face detection. Likewise, the Histogram of Oriented Gradients (HOG)~\cite{lowe1999object} and the Deformable Part-Based Model (DPM)~\cite{felzenszwalb2008discriminatively} extracted features for various object classes. When the AlexNet model won the ImageNet competition, it demonstrated the power of Convolutional Neural Networks (CNNs) in feature extraction, thus prompting a transition from handcrafted features to automated feature learning. Since then, CNN-based methods have become the mainstream in the field of object detection.

Subsequent detection algorithms can be broadly categorized into two main branches: \textbf{two-stage detectors} and \textbf{one-stage detectors}.

\paragraph{Two-Stage Detectors}
The earliest two-stage detector is R-CNN~\cite{uijlings2013selective}, which applies a CNN to extract features and then uses linear SVM classifiers to predict object boundaries and classes. However, this approach suffers from redundant feature computations across numerous overlapping proposals, causing a very slow detection speed. To address this, SPPNet~\cite{he2015spatial} introduced a spatial pyramid pooling (SPP) layer, which avoids redundant computations by pooling features in a multi-scale manner. Although SPPNet runs more than 20 times faster than R-CNN, it still faces challenges such as multi-stage training. Fast R-CNN~\cite{girshick2015fast} further improved efficiency by sharing convolutional layers for region proposals, while Faster R-CNN~\cite{ren2016faster} incorporated a Region Proposal Network (RPN), thereby enabling nearly cost-free region proposals and facilitating an end-to-end learning framework.

\paragraph{One-Stage Detectors}
Representative one-stage detectors include SSD (Single Shot Multibox Detector) and the YOLO (You Only Look Once) series. SSD~\cite{liu2016ssd} employs anchor boxes and multi-scale feature maps to enhance both accuracy and speed. In contrast, YOLO~\cite{redmon2016you} processes the entire image with a single neural network, dividing the image into regions and predicting bounding boxes and class probabilities for each region simultaneously. While YOLO is extremely fast, it typically exhibits slightly lower accuracy than two-stage detectors, particularly for small or heavily occluded objects. Later versions of YOLO~\cite{redmon2018yolov3,bochkovskiy2020yolov4optimalspeedaccuracy,ultralytics2021yolov5} focused on addressing these shortcomings. Notably, Ultralytics~\cite{Jocher_Ultralytics_YOLO_2023} became a key developer and maintainer of the YOLO series from YOLOv5 onward. YOLOv8 introduced significant improvements in architecture, performance, and user experience, further optimizing the YOLO framework for diverse applications such as thermal imaging and real-time edge device deployment. 

Through extensive experiments, we demonstrate that YOLOv8 achieves superior performance in occlusion scenarios on multi-object detection tasks. Therefore, we adopt YOLOv8 as our baseline architecture and further enhance it for occlusion handling and lightweight multi-object detection.

\subsection{Lightweight Models}

Convolutional Neural Network (CNN) methods often entail high computational complexity and large numbers of parameters, presenting substantial hurdles for deploying object detection on edge and mobile devices. Therefore, accelerating detection has emerged as a key objective. One widely adopted approach is network pruning~\cite{cheng2024surveydeepneuralnetwork}, which discards less significant parameters or modules following each training cycle. Alternatively, a more direct strategy involves developing inherently lightweight networks from the outset.

Several noteworthy lightweight networks have been proposed over the years. ShuffleNet~\cite{zhang2017shufflenetextremelyefficientconvolutional} harnesses group convolutions to split feature channels into $m$ groups, potentially reducing computational costs to just $1/m$ of their original level. Building on that, ShuffleNetV2~\cite{ma2018shufflenetv2practicalguidelines} sets forth four guidelines for efficient network design, which include Channel Split to further diminish computational overhead. SqueezeNet~\cite{iandola2016squeezenetalexnetlevelaccuracy50x} lowers costs by shrinking the input layer right from the beginning. Meanwhile, NAS-FCOS~\cite{wang2020nasfcosfastneuralarchitecture} explores a reinforcement learning framework to optimize both the feature pyramid network (FPN) architecture and the prediction head for an anchor-free FCOS detector. GhostNet~\cite{han2020ghostnetfeaturescheapoperations} employs a set of linear transformations that generate numerous “ghost” feature maps, revealing additional insights embedded in intrinsic features.

Another influential family of lightweight networks is the MobileNet series. MobileNet~\cite{howard2017mobilenetsefficientconvolutionalneural} replaces standard convolutions with depthwise separable convolutions, thereby trimming computational demands, and introduces two global hyperparameters to balance latency and accuracy. MobileNetV2~\cite{sandler2019mobilenetv2invertedresidualslinear} brings in inverted residuals and linear bottlenecks, which expand features from low-dimensional representations to higher-dimensional ones for efficiency, while retaining sufficient representational capacity when transitioning back to lower dimensions. MobileNetV3~\cite{howard2019searchingmobilenetv3} employs Neural Architecture Search (NAS) to identify high-performance network structures, along with Squeeze-and-Excitation (SE) modules for enhanced feature representations. It also replaces ReLU with Swish and Hard-Swish to strengthen non-linear expressiveness while moderating computational load. Lastly, MobileNetV4~\cite{qin2024mobilenetv4universalmodels} inherits NAS-based techniques, incorporates a Universal Inverted Bottleneck (UIB), and introduces the Mobile MQA attention module, collectively improving inference speed and efficiency without compromising accuracy. As a result, it achieves near-optimal performance across various computing platforms and AI accelerators.

Despite these advancements, most lightweight frameworks still grapple with sustaining high accuracy under challenging conditions such as occlusion or low-light environments. In our work, we adopt MobileNetV4 to optimize both speed and detection quality, targeting real-time infrared applications on resource-constrained devices.

\subsection{Imbalance Data}

Imbalanced data represents a significant obstacle in multi-label object detection, as real-world datasets are often naturally skewed. When researchers gather data from diverse environments, certain classes tend to dominate, while others remain underrepresented. Although this phenomenon is generally unavoidable, addressing it is essential for achieving robust detection performance. Two main strategies have been widely adopted to mitigate the class imbalance problem: \textbf{resampling} approaches and \textbf{loss-function} based solutions.

\paragraph{Resampling}
Resampling seeks to adjust the class ratio, aiming to create a more balanced distribution among samples. For instance, the K-Nearest Neighbor Oversampling approach (KNNOR)~\cite{ISLAM2022108288} generates synthetic data points for minority classes to compensate for their scarcity. Another technique, the random forest cleaning rule (RFCL)~\cite{Zhang2020RFCLAN}, removes majority-class instances that cross the margin threshold of a newly defined classification boundary, thereby preventing these dominant samples from overwhelming the model.

\paragraph{Loss Function}
An alternative way to address class imbalance is through specialized loss functions. One study~\cite{9324926} proposed a dynamically weighted loss function that rebalances classes by assigning weights based on both class frequency and the predicted probability for the ground-truth class. Another work~\cite{YEUNG2022102026} introduced the Unified Focal Loss, a hierarchical framework that generalizes Dice and cross-entropy-based losses to handle class imbalance, and successfully applied it to medical datasets. Additionally, hard samples—often marked by fewer instances—can benefit from increased emphasis during training. For example, SlideLoss~\cite{yu2022yolofacev2scaleocclusionaware} assigns greater weight to these difficult samples, enabling the model to effectively learn from them despite their limited representation.

Current research on class imbalance primarily focuses on classification tasks, with limited exploration in real-time object detection. Our work fills this gap by further applying SlideLoss to real-time detection scenarios involving occlusion and multi-object challenges.

\section{Proposed Method}\label{sec propsed method}

Before detailing our modifications to YOLOv8, we evaluated multiple YOLO variants (YOLOv5n, YOLOv8n, YOLOv9t, YOLOv10n, and YOLOv11n) to identify an optimal baseline for infrared detection. As shown in Table~\ref{tab:model_comparison}, YOLOv8n exhibited the best balance of accuracy and efficiency, particularly after re-training (\(*\)) on our custom dataset. Therefore, YOLOv8 served as the strongest starting point for further enhancements, leading us to develop our proposed \texttt{MS-YOLO} architecture.

\subsection{Our Model Architecture}

\begin{figure*}[t]
    \centering
    \includegraphics[width=0.9\textwidth]{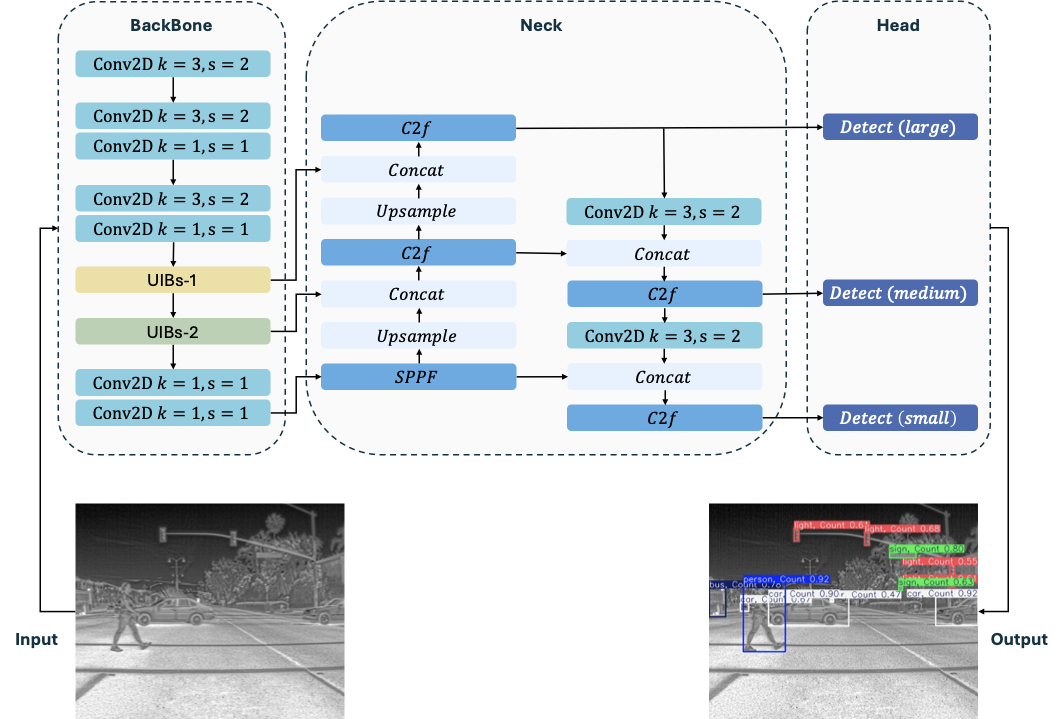}
    \caption{Network architecture of \textbf{MS-YOLO}. \textbf{Conv2D} represents a convolutional operation followed by batch normalization and a ReLU6 activation function. The symbol “\emph{k}” denotes the kernel size, while “\emph{s}” indicates the stride.}
    \label{MNv4_YOLO}
\end{figure*}

YOLOv8, proposed by Ultralytics~\cite{Jocher_Ultralytics_YOLO_2023} in 2023, is a well-known framework among the YOLO family. In this paper, we adopt the YOLOv8 Nano model as our baseline and introduce two major modifications, resulting in our MS-YOLO model as illustrated in Figure~\ref{MNv4_YOLO}. The MS-YOLO model can be divided into three main components: the Backbone, the Neck, and the Head. 

Backbone consists of multiple convolutional blocks, each followed by batch normalization and an activation function. These blocks extract features from the input image (e.g., edges, textures, and complex patterns). As shown in Figure~\ref{MNv4_YOLO}, we replace its original backbone (Darknet-53) with the state-of-the-art lightweight backbone MobileNetv4. This substitution yields faster inference speeds, making the model suitable for edge-device deployment, while it achieves better detection accuracy (see details in Section~\ref{MobileNetv4_Backbone}).

Neck acts as a bridge between the Backbone and the Head by fusing and enhancing feature maps at different scales. This fusion of low-level, mid-level, and high-level features is crucial for accurate object localization and classification in the subsequent stage. We retain the original Neck structure from YOLOv8.

Lastly, our model employs three detection heads to detect objects of various sizes (small, medium, and large). Each detection head processes the corresponding feature map from the Neck to predict bounding boxes, confidence scores, and class labels. We introduce SlideLoss to address multi-label imbalance by leveraging small or hard samples to train the model more effectively (see details in Section~\ref{loss function}).

\subsection{MobileNetv4 Backbone}\label{MobileNetv4_Backbone}

\begin{figure}[htbp]
    \centering
    \includegraphics[width=0.48\textwidth]{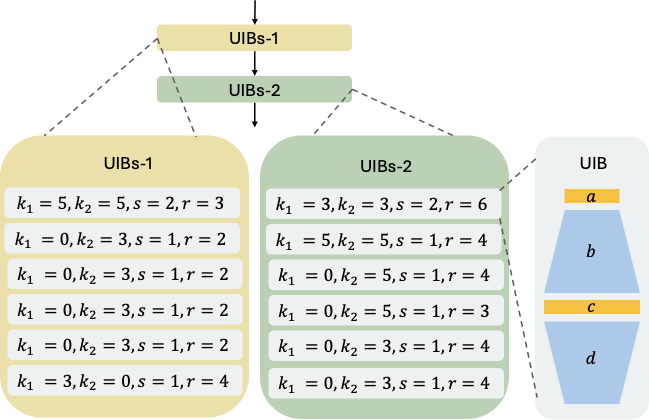}
    \caption{Specifications of UIBs-1 and UIBs-2 in \texttt{MS-YOLO}.}
    \label{UIB_fig}
\end{figure}

Although the YOLOv8 backbone is a high-performing network for general detection tasks, its computational complexity can be prohibitive for edge-device deployment—particularly under the low-light and challenging conditions typical of thermal data. In contrast, \emph{MobileNetV4 Small} offers a more specialized, lightweight architecture. Through aggressive channel reduction and depthwise convolutions, it delivers efficient performance on limited-resource platforms such as thermal sensors or UAV-based surveillance systems.

The backbone begins with a series of Conv2D modules, each comprising a convolution operation followed by batch normalization and ReLU6 activation. Throughout this paper, “\emph{k}” refers to the kernel size, while “\emph{s}” indicates the stride. These initial layers focus on extracting low-level and mid-level features crucial to subsequent detection tasks.

Following these early modules, two UIB (\emph{Universal Inverted Bottleneck}) stack blocks are incorporated to capture more complex patterns. Proposed by Qin et al.~\cite{qin2024mobilenetv4universalmodels}, the UIB framework builds on the Inverted Bottleneck concept from MobileNetV2 but introduces a more modular structure. Each small, LEGO-like block can be easily added or removed by adjusting its parameters, and the overall design is well-matched with Neural Architecture Search (NAS) to systematically identify optimal configurations.

As shown in Figure~\ref{UIB_fig}, our proposed \texttt{MS-YOLO} integrates two types of UIB modules, UIBs-1 and UIBs-2. Each UIB typically comprises:

\begin{enumerate}[label=(\alph*)]
    \item the first depthwise convolution with kernel size $k_1$
    \item a 1$\times$1 expansion convolution
    \item the second depthwise convolution with kernel size $k_2$
    \item a 1$\times$1 projection convolution
\end{enumerate}
The stride $s$ is applied to the middle depthwise layer, while the expansion ratio $r$ governs how many internal channels are used before projecting back down. If $k_1$ or $k_2$ is zero, that depthwise step is skipped entirely, making these blocks highly flexible or ''LEGO-like''. By stacking these lightweight yet efficient blocks, we minimize computational overhead while preserving robust feature extraction capabilities. Consequently, \emph{MobileNetV4 Small} is a more practical choice for embedded and low-power applications requiring rapid inference on complex data.

\subsection{Optimization of Loss Function---SlideLoss}\label{loss function}

YOLOv8's original multi-part loss integrates terms for objectness, bounding box regression, and classification. While this formulation achieves strong overall accuracy~\cite{9042296}, it can struggle with certain classes in multi-label scenarios. In particular, common classes may dominate the training process, overshadowing minority classes and limiting the model’s ability to detect them accurately.

To address this issue, we incorporate \emph{SlideLoss}~\cite{yu2022yolofacev2scaleocclusionaware}, which adaptively increases the emphasis on more challenging samples, often those appearing with lower frequency. Specifically, we define a threshold parameter \(\mu\), based on the average IoU across all bounding boxes in the training set. For each prediction-target pair with IoU \(x\), we assign a weight \(f(x)\) as follows:

\begin{equation}
f(x) = 
\begin{cases}
1, & x \le \mu - 0.1, \\[6pt]
e^{1-\mu}, & \mu -0.1 < x < \mu, \\[6pt]
e^{1-x}, & x \ge \mu.
\end{cases}
\end{equation}

When \(x\) is below or equal to \(\mu - 0.1\), the weight remains at \(1\). For samples whose IoU lies just below \(\mu\), the weight is increased to \(e^{1-\mu}\), reflecting their status as “hard” examples that require more attention. Finally, samples with IoU above \(\mu\) receive exponentially decreasing weights \(e^{1-x}\), ensuring that while the model still learns from well-classified instances, it does not allow these simpler cases to dominate training.

By replacing or augmenting YOLOv8’s original classification and regression losses with SlideLoss, we mitigate the adverse effects of label imbalance. Challenging or underrepresented examples are prioritized, ultimately improving detection performance across all classes without a significant increase in computational overhead.

\section{Results} \label{sec result}

\subsection{Experimental Training Setup}

\paragraph{Dataset} \label{dataset}

We utilize the FLIR ADAS V2~\cite{FLIR2023} dataset, which was published in 2023 and is both new and of high quality. The Teledyne FLIR thermal sensors are particularly advantageous for challenging conditions such as total darkness, fog, smoke, inclement weather, and glare, making them well-suited for the obscured detection tasks in our study. This dataset contains a total of 30{,}787 fully annotated frames, comprising over 520{,}000 bounding box annotations spanning 15 different object categories.

\begin{figure}[htbp]
    \centering
    \includegraphics[width=0.5\textwidth]{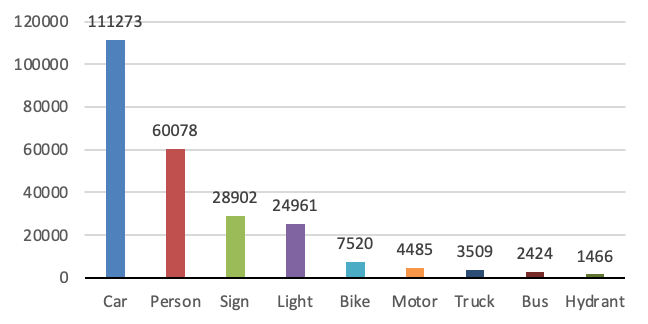}
    \caption{Class Distribution of the Nine Selected Categories in the Thermal Dataset}
    \label{freq}
\end{figure}

From these 15 categories, we select 9 that are commonly encountered in daily life—\textit{person}, \textit{bike}, \textit{car}, \textit{motor}, \textit{bus}, \textit{truck}, \textit{light}, \textit{hydrant}, and \textit{sign}—yielding 15{,}094 images and 244{,}618 object instances. Although the dataset contains both RGB and thermal images, in this study we exclusively use the thermal images. Figure~\ref{freq} shows the class distribution of the selected categories in the dataset.

We focus on these categories not only because they are prevalent in everyday scenarios, but also because their varying sample sizes enable us to investigate performance across both abundant and rare classes. Table~\ref{tab:thermal-stats} summarizes the frequency of each class under thermal modes.

\begin{table}[htbp]
\caption{Dataset Statistics for Thermal Images}
\label{tab:thermal-stats}
\centering
\begin{tabular}{lrrrr}
\toprule
\textbf{Type} & \textbf{Pictures} & \textbf{Instances} & \textbf{Pictures (\%)} & \textbf{Instances (\%)} \\
\midrule
Train      & 10{,}474 & 167{,}640 & 69.39\% & 68.53\% \\
Test       & 3{,}493  & 60{,}515  & 23.14\% & 24.74\% \\
Validation & 1{,}127  & 16{,}463  & 7.47\%  & 6.73\%  \\
\bottomrule
\end{tabular}
\end{table}

Another reason for choosing these labels is that they map directly to classes in the COCO dataset~\cite{lin2014microsoft}, which allows us to leverage pretrained YOLO models. This setup enables a fair evaluation of how well such models---originally trained on COCO---perform on our thermal dataset, and provides a basis for comparison with our own custom-trained models.

All models in our experiments were trained on a Linux-based HPC cluster equipped with a single NVIDIA Tesla V100 SXM2 GPU. Each model was trained for 200 epochs, using an input resolution of \(640 \times 640\) pixels, with a batch size of 128. We employed these consistent settings to facilitate comparisons across different model architectures and training regimes.

\paragraph{Evaluation Criteria}

In this paper, we employ several standard metrics to evaluate both the accuracy and computational efficiency of our object detection model, as described below.

\begin{itemize}

    \begin{figure}[htbp]
    \centering
    \includegraphics[width=0.4\textwidth]{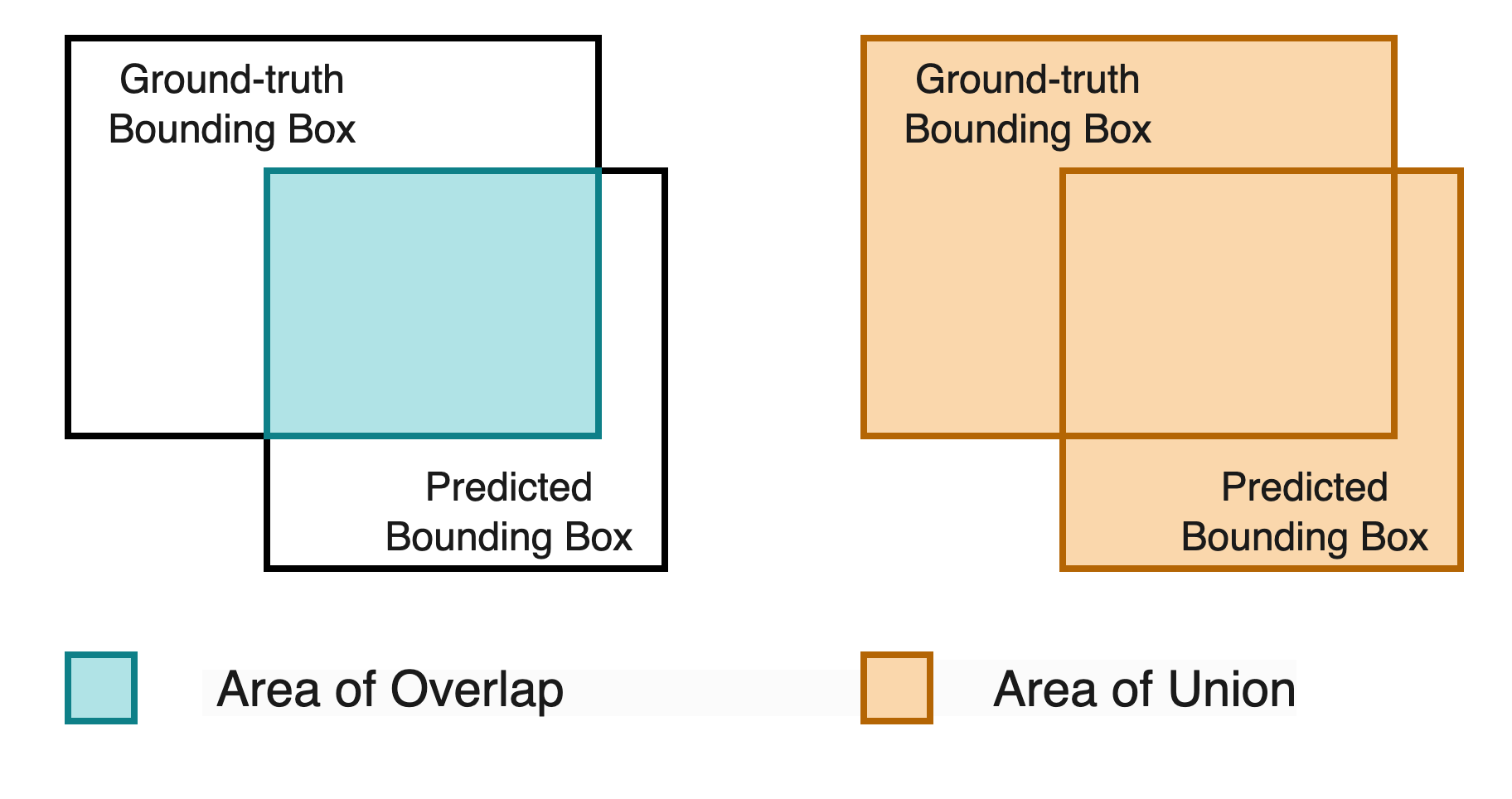}
    \caption{Illustration of the Overlap between a Detected Box and Its Ground-truth.}
    \label{iou}
    \end{figure}

    \item \textbf{Intersection over Union (IoU):}
    IoU quantifies how much a predicted bounding box overlaps with the corresponding ground-truth box, relative to their combined area (see Figure~\ref{iou}). A detection is deemed a true positive if its IoU exceeds the threshold of 0.5 set in this work. Formally:
    \begin{equation}
    \text{IoU} = \frac{\text{Area of Overlap}}{\text{Area of Union}}
    \end{equation}

    \item \textbf{Precision (P):}
    Precision measures how many of the model’s predicted bounding boxes are correct. It is computed as:
    \begin{equation}
    \text{Precision} = \frac{\text{True Positives}}{\text{True Positives} + \text{False Positives}}
    \end{equation}

    \item \textbf{Recall (R):}
    Recall indicates the proportion of ground-truth bounding boxes that the model successfully detects. It is given by:
    \begin{equation}
    \text{Recall} = \frac{\text{True Positives}}{\text{True Positives} + \text{False Negatives}}
    \end{equation}

    \item \textbf{Average Precision (AP):}
    AP is defined as the area under the Precision–Recall curve for a particular class, effectively capturing the balance between these two metrics:
    \begin{equation}
    \text{AP} = \int_{0}^{1} \text{Precision}(r) \, dr
    \end{equation}
    where \(r\) represents the recall value.

    \item \textbf{Mean Average Precision (mAP):}
    mAP offers an aggregate measure of detection performance across all classes. It is the average of the AP values for each class:
    \begin{equation}
    \text{mAP} = \frac{1}{M} \sum_{i=1}^{M} \text{AP}_i
    \end{equation}
    where \(M\) denotes the total number of classes.

    \item \textbf{Giga Floating Point Operations (GFLOPs):}
    Gflops represent the number of billions of floating-point operations the model performs to process a single input. A lower Gflops value typically indicates faster inference and lower power consumption, making the model more suitable for resource-constrained environments.

\end{itemize}

\subsection{Ablation Study}

The ablation study presented in Table \ref{tab:ablation_study} evaluates the impact of the MobileNetv4 and SlideLoss modules on model performance. Three configurations were tested: MobileNetv4 only, SlideLoss only, and the combination of both modules.

\begin{table}[htbp]
\centering
\setlength{\tabcolsep}{2pt} 
\caption{Ablation Study of MobileNetv4 and SlideLoss Modules}
\label{tab:ablation_study}
\begin{tabular}{cccccccc}
\toprule
\textbf{\#} & \textbf{MobileNetv4} & \textbf{SlideLoss} & \textbf{Precision} & \textbf{Recall} & \textbf{mAP50} & \textbf{mAP50-95} & \textbf{Gflops} \\
\midrule
1  & - & - & 0.624 & 0.463 & 0.517 & 0.318 & 6.8 \\
2  & \checkmark & - & 0.643 & 0.426 & 0.477 & 0.282 & 6.7 \\
3  & - & \checkmark & 0.636 & 0.459 & 0.505 & 0.300 & 6.8 \\
4  & \checkmark & \checkmark & 0.649 & 0.430 & 0.484 & 0.282 & 6.7 \\
\bottomrule
\end{tabular}
\end{table}

When the MobileNetv4 module is used alone, the model achieves a Precision of 0.643, Recall of 0.426, mAP50 of 0.477, and mAP50-95 of 0.282, with a computational cost of 6.7 Gflops. This reduction of 0.1 GFLOPs (from 6.8 GFLOPs to 6.7 GFLOPs) translates to an approximately 1.5\% increase in inference speed. These results indicate that our model requires fewer computational resources during inference while maintaining solid performance, demonstrating the effectiveness of the MobileNetv4 module in maintaining a balance between detection accuracy and computational efficiency. 

The SlideLoss module alone improves Recall to 0.459 and achieves the highest mAP50 and mAP50-95 values of 0.505 and 0.300, respectively. However, this comes at a slightly higher computational cost of 6.8 Gflops, highlighting its strength in improving model generalization and accuracy at the cost of efficiency.

When both modules are combined, the model achieves the highest Precision (0.649) and Recall (0.430) while maintaining competitive mAP50 (0.484) and mAP50-95 (0.282) scores. Importantly, the computational cost remains efficient at 6.7 Gflops. This configuration demonstrates the complementary benefits of integrating MobileNetv4 and SlideLoss modules, achieving a balance of accuracy and efficiency, making it the optimal configuration for our model.

\subsection{Comparison Experiments}

\paragraph{Model Overview}
Table~\ref{tab:model_comparison} presents performance results for various YOLO-based models. The “Nano” and “Tiny” designations (e.g., YOLOv5n, YOLOv9t) refer to smaller architectures, while the asterisk (\(*\)) indicates versions that have been re-trained on our custom dataset described in Section~\ref{dataset}. Our proposed model, \texttt{MS-YOLO}, employs a MobileNetv4-based backbone with SlideLoss, combining lightweight design and adaptive loss to tackle class imbalance.

\begin{table}[htbp]
\centering
\setlength{\tabcolsep}{5.5pt} 
\caption{Model Performance Comparison}
\label{tab:model_comparison}
\begin{tabular}{lccccc}
\toprule
\textbf{Model name} & \textbf{Precision} & \textbf{Recall} & \textbf{mAP50} & \textbf{mAP50-95} & \textbf{Gflops} \\
\midrule
YOLOv5n          & 0.415 & 0.154 & 0.167 & 0.099 & 7.7 \\
YOLOv8n          & 0.426 & 0.171 & 0.181 & 0.106 & 8.7 \\
YOLOv9t          & 0.399 & 0.199 & 0.212 & 0.125 & 8.2 \\
YOLOv10n         & 0.353 & 0.173 & 0.179 & 0.107 & 8.6 \\
YOLOv11n         & 0.375 & 0.182 & 0.189 & 0.109 & 6.5 \\
YOLOv5n*   & 0.619 & 0.413 & 0.478 & 0.288 & 5.8 \\
YOLOv8n*  & 0.624 & 0.463 & 0.517 & 0.318 & 6.8 \\
YOLOv9t*   & 0.628 & 0.418 & 0.462 & 0.274 & 8.2 \\
YOLOv10n*  & 0.593 & 0.422 & 0.477 & 0.281 & 8.2 \\
YOLOv11n*  & 0.628 & 0.418 & 0.462 & 0.274 & 8.2 \\
\textbf{MS-YOLO} & \textbf{0.649} & \textbf{0.430} & \textbf{0.484} & \textbf{0.282} & \textbf{6.7} \\
\bottomrule
\end{tabular}
\end{table}

\paragraph{Accuracy Metrics}
In terms of detection quality, precision and recall reveal trade-offs in how models identify and correctly predict objects. Across original YOLO variants (YOLOv5n - YOLOv11n), performance is moderate, often due to limited adaptation to our specific dataset. However, retraining these models (\(*\) versions) significantly boosts their capabilities. Among all methods, \texttt{MS-YOLO} achieves the highest precision of 0.649, closely followed by YOLOv9t* and YOLOv11n* at around 0.628. While YOLOv8n* stands out with the highest recall (0.463) and leads in mAP50 (0.517) and mAP50-95 (0.318), \texttt{MS-YOLO} also delivers a robust performance with 0.430 recall, 0.484 mAP50, and 0.282 mAP50-95.

\paragraph{Computational cost}
Speed and computational cost are measured in Gflops. Models such as YOLOv5n* have the lowest demand at 5.8 Gflops, while YOLOv8n consumes 8.7 GFLOPs, reflecting its high computational cost. Our \texttt{MS-YOLO} operates at 6.7 Gflops, striking a balance between efficiency and accuracy. This moderate resource requirement allows deployment in settings where real-time inference and limited computational power are both critical.

\paragraph{Summary}
Overall, \texttt{MS-YOLO} is a strong model in scenarios demanding both reliable detection quality and manageable computational overhead. By integrating a MobileNetv4-based backbone and SlideLoss, it achieves high precision, competitive recall, and robust mAP performance. Coupled with a relatively small Gflops, it stands out as a promising solution for resource-constrained environments.

\subsection{Class-Level Insights and Imbalance Analysis}

Table~\ref{tab:class_performance} outlines the class-level performance of our \texttt{MS-YOLO} model. While these metrics indicate satisfactory results overall, there is notable variation among individual classes. In general, classes with higher representation tend to achieve stronger performance. For example, \emph{person} (4309 instances) and \emph{car} (7128 instances) exhibit robust detection results, demonstrating that larger sample sizes often aid in more effective feature learning.

\begin{table}[htbp]
\centering
\caption{Performance Evaluation of MS-YOLO Across Classes}
\label{tab:class_performance}
\small
\setlength{\tabcolsep}{1.1pt} 
\begin{tabular}{lccccccc}
\toprule
\textbf{Class} & \textbf{Pictures} & \textbf{Instances} & \textbf{Precision} & \textbf{Recall} & \textbf{mAP50} & \textbf{mAP50-95} \\
\midrule
all             & 1144 & 16455 & 0.649 & 0.430 & 0.484 & 0.282 \\
person          & 791  & 4309  & 0.803 & 0.586 & 0.714 & 0.386 \\
bicycle         & 135  & 170   & 0.414 & 0.453 & 0.439 & 0.260 \\
car             & 1047 & 7128  & 0.810 & 0.729 & 0.814 & 0.570 \\
motorcycle      & 47   & 55    & 0.761 & 0.579 & 0.607 & 0.297 \\
bus             & 126  & 179   & 0.780 & 0.385 & 0.548 & 0.370 \\
truck           & 42   & 46    & 0.191 & 0.261 & 0.0965 & 0.0652 \\
traffic light   & 439  & 2002  & 0.703 & 0.345 & 0.432 & 0.189 \\
fire hydrant    & 93   & 94    & 0.676 & 0.177 & 0.264 & 0.145 \\
stop sign       & 848  & 2472  & 0.703 & 0.352 & 0.438 & 0.255 \\
\bottomrule
\end{tabular}
% }
\end{table}

However, the relationship between instance count and performance is not always straightforward. While some classes with fewer instances, such as \emph{truck}, show poor performance in both precision (0.191) and recall (0.261), others like \emph{fire hydrant} (94 instances in the validation set) achieve relatively high precision (0.676). This suggests that factors beyond data quantity, such as visual similarity between classes, also play a significant role. For instance, \emph{truck} may be confused with \emph{car} due to their similar appearances, further complicating detection.

%In summary, while data distribution differences across classes are essential， we must also pay attention to the inherent visual and structural differences between classes. These factors, such as shape, size, and texture, can significantly impact model performance and should be carefully considered.

In summary, while class distribution is essential, inherent visual and structural differences—such as shape, size, and texture—also significantly impact model performance and must be carefully considered.

\section{Conclusion} \label{sec conclusion}

This study highlighted two main barriers in infrared urban detection: computational complexity and class imbalance. By benchmarking various YOLO models, we identified YOLOv8 as a strong baseline for its balanced speed and accuracy. Building on this foundation, we integrated MobileNetV4 to reduce inference overhead and ensure feasibility for edge devices, such as drones or embedded traffic systems. To tackle imbalance-related challenges, particularly for occluded or minority-class objects, we introduced SlideLoss, an adaptive loss function that strategically emphasizes hard-to-detect targets.

Our experiments on the FLIR ADAS V2 dataset confirmed that \texttt{MS-YOLO} outperforms conventional architectures in both precision and recall while maintaining low computational demands. This makes it a scalable and efficient solution for real-world applications, particularly in edge devices like drones, smart surveillance, and autonomous urban systems operating under low-light and occlusion-heavy conditions. 

Despite these advancements, the impact of data distribution and inter-class similarities on real-time infrared object detection remains an open question. Further research is needed to refine adaptive loss functions and explore cross-domain learning approaches that enhance generalization in complex urban environments.

\section*{Acknowledgment}
This work was supported by the Air Force Research Laboratory (AFRL) and the Lifelong Learning Machines program by DARPA/MTO under Contract No. FA8650-18-C-7831. Any opinions, findings, conclusions, or recommendations expressed in this material are those of the author(s) and do not necessarily reflect the views of these agencies. The research was also sponsored by the Army Research Laboratory and was accomplished under Cooperative Agreement Number W911NF-22-2-0209. The views and conclusions contained in this document are those of the authors and should not be interpreted as representing the official policies, either expressed or implied, of the Army Research Laboratory or the U.S. Government. The U.S. Government is authorized to reproduce and distribute reprints for Government purposes, notwithstanding any copyright notation herein. This work was also supported in part by the Simons Collaboration Grant No. 711999.

\bibliographystyle{IEEEtran}
\bibliography{refs}

\begin{thebibliography}{10}
\providecommand{\url}[1]{#1}
\csname url@samestyle\endcsname
\providecommand{\newblock}{\relax}
\providecommand{\bibinfo}[2]{#2}
\providecommand{\BIBentrySTDinterwordspacing}{\spaceskip=0pt\relax}
\providecommand{\BIBentryALTinterwordstretchfactor}{4}
\providecommand{\BIBentryALTinterwordspacing}{\spaceskip=\fontdimen2\font plus
\BIBentryALTinterwordstretchfactor\fontdimen3\font minus \fontdimen4\font\relax}
\providecommand{\BIBforeignlanguage}[2]{{%
\expandafter\ifx\csname l@#1\endcsname\relax
\typeout{** WARNING: IEEEtran.bst: No hyphenation pattern has been}%
\typeout{** loaded for the language `#1'. Using the pattern for}%
\typeout{** the default language instead.}%
\else
\language=\csname l@#1\endcsname
\fi
#2}}
\providecommand{\BIBdecl}{\relax}
\BIBdecl

\bibitem{zheng2023nighttime}
Q.~Zheng, K.~C. Seto, Y.~Zhou, S.~You, and Q.~Weng, ``Nighttime light remote sensing for urban applications: Progress, challenges, and prospects,'' \emph{ISPRS Journal of Photogrammetry and Remote Sensing}, vol. 202, pp. 125--141, 2023.

\bibitem{shyam2024lightweight}
P.~Shyam and H.~Yoo, ``Lightweight thermal super-resolution and object detection for robust perception in adverse weather conditions,'' in \emph{Proceedings of the IEEE/CVF Winter Conference on Applications of Computer Vision}, 2024, pp. 7471--7482.

\bibitem{collini2024flexible}
E.~Collini, L.~A.~I. Palesi, P.~Nesi, G.~Pantaleo, and W.~Zhao, ``Flexible thermal camera solution for smart city people detection and counting,'' \emph{Multimedia Tools and Applications}, vol.~83, no.~7, pp. 20\,457--20\,485, 2024.

\bibitem{eltahan2024enhancing}
M.~Eltahan and K.~Elsayed, ``Enhancing autonomous driving by exploiting thermal object detection through feature fusion,'' \emph{International Journal of Intelligent Transportation Systems Research}, vol.~22, no.~1, pp. 146--158, 2024.

\bibitem{Jocher_Ultralytics_YOLO_2023}
\BIBentryALTinterwordspacing
G.~Jocher, J.~Qiu, and A.~Chaurasia, ``{Ultralytics YOLO},'' Jan. 2023. [Online]. Available: \url{https://github.com/ultralytics/ultralytics}
\BIBentrySTDinterwordspacing

\bibitem{tan2019efficientnet}
M.~Tan and Q.~Le, ``Efficientnet: Rethinking model scaling for convolutional neural networks,'' in \emph{International conference on machine learning}.\hskip 1em plus 0.5em minus 0.4em\relax PMLR, 2019, pp. 6105--6114.

\bibitem{chen2024modeling}
X.~Chen, L.~Chen, L.~Chen, P.~Chen, G.~Sheng, X.~Yu, and Y.~Zou, ``Modeling thermal infrared image degradation and real-world super-resolution under background thermal noise and streak interference,'' \emph{IEEE Transactions on Circuits and Systems for Video Technology}, vol.~33, no.~9, pp. 4496 -- 4506, 2024.

\bibitem{9042296}
K.~Oksuz, B.~C. Cam, S.~Kalkan, and E.~Akbas, ``Imbalance problems in object detection: A review,'' \emph{IEEE Transactions on Pattern Analysis and Machine Intelligence}, vol.~43, no.~10, pp. 3388--3415, 2021.

\bibitem{ren2016faster}
S.~Ren, K.~He, R.~Girshick, and J.~Sun, ``Faster r-cnn: Towards real-time object detection with region proposal networks,'' \emph{IEEE transactions on pattern analysis and machine intelligence}, vol.~39, no.~6, pp. 1137--1149, 2016.

\bibitem{mittal2024comprehensive}
P.~Mittal, ``A comprehensive survey of deep learning-based lightweight object detection models for edge devices,'' \emph{Artificial Intelligence Review}, vol.~57, no.~9, p. 242, 2024.

\bibitem{qin2024mobilenetv4universalmodels}
\BIBentryALTinterwordspacing
D.~Qin, C.~Leichner, M.~Delakis, M.~Fornoni, S.~Luo, F.~Yang, W.~Wang, C.~Banbury, C.~Ye, B.~Akin, V.~Aggarwal, T.~Zhu, D.~Moro, and A.~Howard, ``Mobilenetv4 -- universal models for the mobile ecosystem,'' 2024. [Online]. Available: \url{https://arxiv.org/abs/2404.10518}
\BIBentrySTDinterwordspacing

\bibitem{yu2022yolofacev2scaleocclusionaware}
\BIBentryALTinterwordspacing
Z.~Yu, H.~Huang, W.~Chen, Y.~Su, Y.~Liu, and X.~Wang, ``Yolo-facev2: A scale and occlusion aware face detector,'' 2022. [Online]. Available: \url{https://arxiv.org/abs/2208.02019}
\BIBentrySTDinterwordspacing

\bibitem{FLIR2023}
T.~FLIR, ``Free teledyne flir thermal dataset for algorithm training,'' \url{https://www.flir.com/oem/adas/adas-dataset-form/}, 2023, accessed: 2024-01-30.

\bibitem{viola2001rapid}
P.~Viola and M.~Jones, ``Rapid object detection using a boosted cascade of simple features,'' in \emph{Proceedings of the 2001 IEEE Computer Society Conference on Computer Vision and Pattern Recognition (CVPR 2001)}, vol.~1.\hskip 1em plus 0.5em minus 0.4em\relax Kauai, HI, USA: IEEE, Dec. 2001, pp. 511--518.

\bibitem{lowe1999object}
D.~G. Lowe, ``Object recognition from local scale-invariant features,'' in \emph{Proceedings of the seventh IEEE international conference on computer vision}, vol.~2.\hskip 1em plus 0.5em minus 0.4em\relax Ieee, 1999, pp. 1150--1157.

\bibitem{felzenszwalb2008discriminatively}
P.~Felzenszwalb, D.~McAllester, and D.~Ramanan, ``A discriminatively trained, multiscale, deformable part model,'' in \emph{2008 IEEE conference on computer vision and pattern recognition}.\hskip 1em plus 0.5em minus 0.4em\relax Ieee, 2008, pp. 1--8.

\bibitem{uijlings2013selective}
J.~R. Uijlings, K.~E. Van De~Sande, T.~Gevers, and A.~W. Smeulders, ``Selective search for object recognition,'' \emph{International journal of computer vision}, vol. 104, pp. 154--171, 2013.

\bibitem{he2015spatial}
K.~He, X.~Zhang, S.~Ren, and J.~Sun, ``Spatial pyramid pooling in deep convolutional networks for visual recognition,'' \emph{IEEE transactions on pattern analysis and machine intelligence}, vol.~37, no.~9, pp. 1904--1916, 2015.

\bibitem{girshick2015fast}
R.~Girshick, ``Fast r-cnn,'' \emph{arXiv preprint arXiv:1504.08083}, 2015.

\bibitem{liu2016ssd}
W.~Liu, D.~Anguelov, D.~Erhan, C.~Szegedy, S.~Reed, C.-Y. Fu, and A.~C. Berg, ``Ssd: Single shot multibox detector,'' in \emph{Computer Vision--ECCV 2016}.\hskip 1em plus 0.5em minus 0.4em\relax Springer, 2016, pp. 21--37.

\bibitem{redmon2016you}
J.~Redmon, S.~Divvala, R.~Girshick, and A.~Farhadi, ``You only look once: Unified, real-time object detection,'' in \emph{Proceedings of the 2016 IEEE Conference on Computer Vision and Pattern Recognition (CVPR)}.\hskip 1em plus 0.5em minus 0.4em\relax IEEE, 2016, pp. 779--788.

\bibitem{redmon2018yolov3}
J.~Redmon, ``Yolov3: An incremental improvement,'' \emph{arXiv preprint arXiv:1804.02767}, 2018.

\bibitem{bochkovskiy2020yolov4optimalspeedaccuracy}
\BIBentryALTinterwordspacing
A.~Bochkovskiy, C.-Y. Wang, and H.-Y.~M. Liao, ``Yolov4: Optimal speed and accuracy of object detection,'' 2020. [Online]. Available: \url{https://arxiv.org/abs/2004.10934}
\BIBentrySTDinterwordspacing

\bibitem{ultralytics2021yolov5}
Ultralytics, ``{YOLOv5}: {A} state-of-the-art real-time object detection system,'' \url{https://docs.ultralytics.com}, 2021, accessed: insert date here.

\bibitem{cheng2024surveydeepneuralnetwork}
\BIBentryALTinterwordspacing
H.~Cheng, M.~Zhang, and J.~Q. Shi, ``A survey on deep neural network pruning-taxonomy, comparison, analysis, and recommendations,'' 2024. [Online]. Available: \url{https://arxiv.org/abs/2308.06767}
\BIBentrySTDinterwordspacing

\bibitem{zhang2017shufflenetextremelyefficientconvolutional}
\BIBentryALTinterwordspacing
X.~Zhang, X.~Zhou, M.~Lin, and J.~Sun, ``Shufflenet: An extremely efficient convolutional neural network for mobile devices,'' 2017. [Online]. Available: \url{https://arxiv.org/abs/1707.01083}
\BIBentrySTDinterwordspacing

\bibitem{ma2018shufflenetv2practicalguidelines}
N.~Ma, X.~Zhang, H.-T. Zheng, and J.~Sun, ``Shufflenet v2: Practical guidelines for efficient cnn architecture design,'' in \emph{Computer Vision – ECCV 2018}, Sep. 2018, pp. 122--138.

\bibitem{iandola2016squeezenetalexnetlevelaccuracy50x}
\BIBentryALTinterwordspacing
F.~N. Iandola, S.~Han, M.~W. Moskewicz, K.~Ashraf, W.~J. Dally, and K.~Keutzer, ``Squeezenet: Alexnet-level accuracy with 50x fewer parameters and <0.5mb model size,'' 2016. [Online]. Available: \url{https://arxiv.org/abs/1602.07360}
\BIBentrySTDinterwordspacing

\bibitem{wang2020nasfcosfastneuralarchitecture}
\BIBentryALTinterwordspacing
N.~Wang, Y.~Gao, H.~Chen, P.~Wang, Z.~Tian, C.~Shen, and Y.~Zhang, ``Nas-fcos: Fast neural architecture search for object detection,'' 2020. [Online]. Available: \url{https://arxiv.org/abs/1906.04423}
\BIBentrySTDinterwordspacing

\bibitem{han2020ghostnetfeaturescheapoperations}
\BIBentryALTinterwordspacing
K.~Han, Y.~Wang, Q.~Tian, J.~Guo, C.~Xu, and C.~Xu, ``Ghostnet: More features from cheap operations,'' 2020. [Online]. Available: \url{https://arxiv.org/abs/1911.11907}
\BIBentrySTDinterwordspacing

\bibitem{howard2017mobilenetsefficientconvolutionalneural}
\BIBentryALTinterwordspacing
A.~G. Howard, M.~Zhu, B.~Chen, D.~Kalenichenko, W.~Wang, T.~Weyand, M.~Andreetto, and H.~Adam, ``Mobilenets: Efficient convolutional neural networks for mobile vision applications,'' 2017. [Online]. Available: \url{https://arxiv.org/abs/1704.04861}
\BIBentrySTDinterwordspacing

\bibitem{sandler2019mobilenetv2invertedresidualslinear}
\BIBentryALTinterwordspacing
M.~Sandler, A.~Howard, M.~Zhu, A.~Zhmoginov, and L.-C. Chen, ``Mobilenetv2: Inverted residuals and linear bottlenecks,'' 2019. [Online]. Available: \url{https://arxiv.org/abs/1801.04381}
\BIBentrySTDinterwordspacing

\bibitem{howard2019searchingmobilenetv3}
\BIBentryALTinterwordspacing
A.~Howard, M.~Sandler, G.~Chu, L.-C. Chen, B.~Chen, M.~Tan, W.~Wang, Y.~Zhu, R.~Pang, V.~Vasudevan, Q.~V. Le, and H.~Adam, ``Searching for mobilenetv3,'' 2019. [Online]. Available: \url{https://arxiv.org/abs/1905.02244}
\BIBentrySTDinterwordspacing

\bibitem{ISLAM2022108288}
\BIBentryALTinterwordspacing
A.~Islam, S.~B. Belhaouari, A.~U. Rehman, and H.~Bensmail, ``Knnor: An oversampling technique for imbalanced datasets,'' \emph{Applied Soft Computing}, vol. 115, p. 108288, 2022. [Online]. Available: \url{https://www.sciencedirect.com/science/article/pii/S1568494621010942}
\BIBentrySTDinterwordspacing

\bibitem{Zhang2020RFCLAN}
\BIBentryALTinterwordspacing
R.~Zhang, Z.~Zhang, and D.~Wang, ``Rfcl: A new under-sampling method of reducing the degree of imbalance and overlap,'' \emph{Pattern Analysis and Applications}, vol.~24, pp. 641 -- 654, 2020. [Online]. Available: \url{https://api.semanticscholar.org/CorpusID:228879128}
\BIBentrySTDinterwordspacing

\bibitem{9324926}
K.~R.~M. Fernando and C.~P. Tsokos, ``Dynamically weighted balanced loss: Class imbalanced learning and confidence calibration of deep neural networks,'' \emph{IEEE Transactions on Neural Networks and Learning Systems}, vol.~33, no.~7, pp. 2940--2951, 2022.

\bibitem{YEUNG2022102026}
\BIBentryALTinterwordspacing
M.~Yeung, E.~Sala, C.-B. Schönlieb, and L.~Rundo, ``Unified focal loss: Generalising dice and cross entropy-based losses to handle class imbalanced medical image segmentation,'' \emph{Computerized Medical Imaging and Graphics}, vol.~95, p. 102026, 2022. [Online]. Available: \url{https://www.sciencedirect.com/science/article/pii/S0895611121001750}
\BIBentrySTDinterwordspacing

\bibitem{lin2014microsoft}
T.-Y. Lin, M.~Maire, S.~Belongie, J.~Hays, P.~Perona, D.~Ramanan, P.~Doll{\'a}r, and C.~L. Zitnick, ``Microsoft coco: Common objects in context,'' in \emph{Computer Vision--ECCV 2014: 13th European Conference, Zurich, Switzerland, September 6-12, 2014, Proceedings, Part V 13}.\hskip 1em plus 0.5em minus 0.4em\relax Springer, 2014, pp. 740--755.

\end{thebibliography}

\end{document}